\documentclass[runningheads]{llncs}
\usepackage{graphicx}
\usepackage{colortbl}
\usepackage{multirow}%
\usepackage{amsmath,amssymb,amsfonts}%

\usepackage{mathrsfs}%
\usepackage[title]{appendix}%
\usepackage{xcolor}%
\usepackage{textcomp}%
\usepackage{manyfoot}%
\usepackage{booktabs}%
\usepackage{algorithm}%
\usepackage{algorithmicx}%
\usepackage{algpseudocode}%
\usepackage{listings}%
\usepackage[capitalize]{cleveref}
\usepackage[misc]{ifsym}
\begin{document}
\title{UniPose: Unified Cross-modality Pose Prior Propagation towards RGB-D data for Weakly Supervised 3D Human Pose Estimation.}
\titlerunning{UniPose}
\authorrunning{Zheng et al.}
%
\author{Jinghong Zheng \and
Changlong Jiang$^{(\textrm{\Letter})}$ \and
Jiaqi Li \and
Haohong Kuang \and
Hang Xu \and
Tingbing Yan 
}

\institute{National Key Laboratory of Multispectral Information Intelligent Processing Technology, School of Artificial Intelligence and Automation, Huazhong University of Science and Technology, Wuhan 430074, China \\
\email{\{deepzheng, changlongj, lijiaqi\_mail, haohong\_kuang,\\hang\_xu, yantingbing\}@hust.edu.cn}\\
}

%
%

%
\maketitle              
\begin{abstract}

In this paper, we present UniPose, a unified cross-modality pose prior propagation method for weakly supervised 3D human pose estimation (HPE) using unannotated single-view RGB-D sequences (RGB, depth, and point cloud data). UniPose transfers 2D HPE annotations from large-scale RGB datasets (e.g., MS COCO) to the 3D domain via self-supervised learning on easily acquired RGB-D sequences, eliminating the need for labor-intensive 3D keypoint annotations. This approach bridges the gap between 2D and 3D domains without suffering from issues related to multi-view camera calibration or synthetic-to-real data shifts.
During training, UniPose leverages off-the-shelf 2D pose estimations as weak supervision for point cloud networks, incorporating spatial-temporal constraints like body symmetry and joint motion. The 2D-to-3D back-projection loss and cross-modality interaction further enhance this process. By treating the point cloud network's 3D HPE results as pseudo ground truth, our anchor-to-joint prediction method performs 3D lifting on RGB and depth networks, making it more robust against inaccuracies in 2D HPE results compared to state-of-the-art methods.
Experiments on CMU Panoptic and ITOP datasets show that UniPose achieves comparable performance to fully supervised methods. Incorporating large-scale unlabeled data (\textit{e.g.}, NTU RGB+D 60) enhances its performance under challenging conditions, demonstrating its potential for practical applications. Our proposed 3D lifting method also achieves state-of-the-art results.
\end{abstract}

\begin{figure}[h]
\centering
\includegraphics[width=0.92\textwidth]{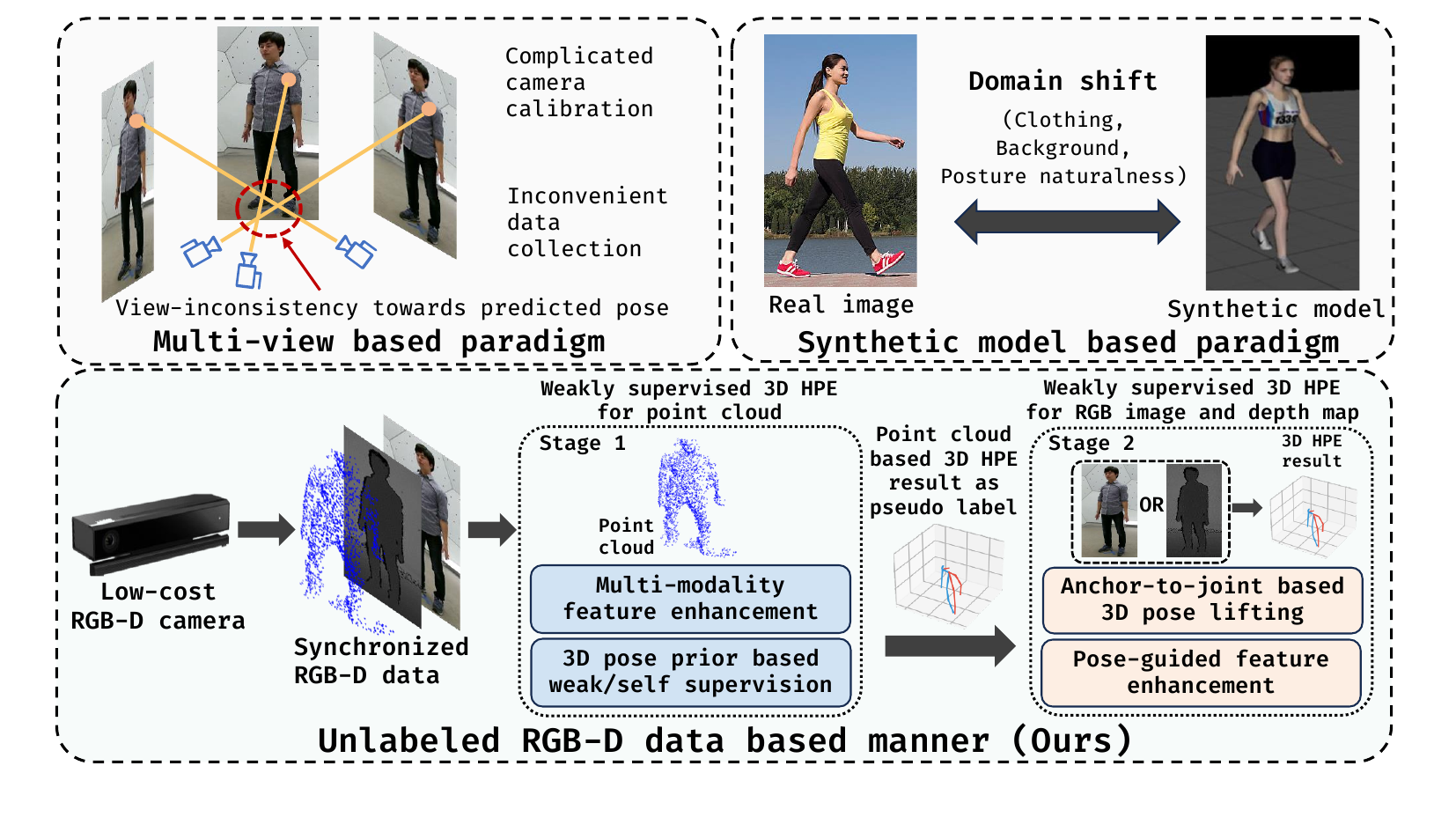}
\vspace{-20pt}
\caption{Comparison among the different weakly supervised 3D human pose estimation manners. Compared with mainstream methods using multi-view images or synthetic models, our proposition does not suffer from the issues of complicated multi-view calibration, inconvenient data collection, and domain shift problem. Within our proposition, RGB data mainly provides weak 3D human pose supervision (i.e., 2D human pose prior), and the paired depth data helps to exploit self 3D human pose supervision. }\label{fig1}
\end{figure}

\section{Introduction}
3D human pose estimation (HPE) based on RGB image, depth map or point cloud, is of wide-range applications towards human action recognition~\cite{liu2017skeleton,liu2017global}, human-robot interaction~\cite{svenstrup2009pose}, virtual reality~\cite{obdrvzalek2012real}, \textit{etc.} 
With the introdution of deep learning technologies~\cite{he2016deep,vaswani2017attention}, 3D HPE's performance has been enhanced remarkably in fully supervised learning manner. 
However, deep network's data-hungry property leads to the high demand on 3D pose annotation both on quality and quantity, which is essentially labor and time consuming. 
As a result, the existing annotated 3D HPE datasets~\cite{Joo_2017_TPAMI} are generally collected under constrained laboratory setting, with indoor environments and small number of subjects performing limited activities.   
Facing challenges of 3D HPE data collection and annotation, leveraging unlabeled data via weakly supervised learning has emerged as an important research task~\cite{cai2018weakly}. This paradigm generally involves assigning noisy pseudo-labels and designing tailored loss functions to distill pose prior from unlabeled data.
Current methods primarily include multi-view setups~\cite{kocabas2019self,mitra2020multiview} and synthetic models~\cite{shrivastava2017learning,cai2018weakly}. Multi-view approaches require complex camera calibration and multiple perspectives, often leading to view-inconsistency and error accumulation when converting 2D poses into 3D via triangulation. Synthetic methods, in another way, suffer from notable domain shift issues between synthetic and real data.
Concerning the essential defects of existing paradigms above, \emph{we propose to leverage weakly supervised 3D HPE task without relying on any prior from multi-view imaging manner or synthetic model}.

Accordingly, we propose a novel weakly supervised 3D HPE approach termed UniPose. 
Compared with the existing weakly supervised counterparts~\cite{shrivastava2017learning,kocabas2019self}, UniPose does not suffer from the complicated multi-view calibration, inconvenient data collection, and domain shift issue between real and synthetic data, due to the introduction of real monocular RGB-D sequence for distilling weak and self supervision clues for 3D HPE. Meanwhile, UniPose's supervision can be applied to the modalities of RGB image, depth map and point cloud uniformly for weakly supervised 3D HPE. 

Acquiring effective supervision signals is one of the crucial problems in weakly supervised 3D HPE. 
We exploit supervision clues from unlabeled RGB-D sequence mainly in terms of weak supervision (e.g. 2D pose prior and correspondence between 2D and 3D pose) and self supervision (e.g., human body physical constraints and motion characteristics).
Towards weak supervision, 2D human pose prior is first acquired from well established 2D HPE model~\cite{sun2019deep} pre-trained on large-scale 2D RGB dataset. 
These 2D HPE results are back-projected into 3D space, minimizing the distance between predicted 3D joints and their corresponding 2D directional projection rays to establish 2D-to-3D correspondences. Since alleviating the dimension compression problem suffered by the existing methods\cite{chen2019weakly} that project 3D joint into 2D space, our approach generally can better reveal 3D characteristics and spatial consistency in 3D space.
We extract self-supervised cues related to real spatial-temporal physical laws of human joints from point cloud sequences. 
In addition to supervision signals, enhancing the representation of point cloud features is also important for accurate 3D HPE. This is achieved by constructing cross-modal interactions, incorporating intermediate features from a pre-trained 2D pose estimator and fusing them with RGB and depth data. This approach not only boosts the 3D descriptive ability of point cloud features but also improves the accuracy of identifying extremity joints like hands and feet. 

With the pseudo 3D HPE annotation generated via point cloud branch, a novel 2D-to-3D pose lifting method is further proposed towards RGB image and depth map for 3D HPE, under anchor-to-joint paradigm~\cite{xiong2019a2j}. For RGB image and depth map, 2D pose is first acquired using well-established 2D HPE methods~\cite{sun2019deep}. Then, the 2D HPE results are lifted to 3D space with our proposed 3D lifting approach, under the supervision of point cloud's pseudo 3D pose annotation. 
Compared with the SOTA 3D pose lifting methods~\cite{zhao2023contextaware,zhang2022mixste} that are generally sensitive to inaccurate 2D HPE results and with relatively high demand on long pose sequence input for generalization, our proposition alleviates this via introducing anchor-to-joint manner with joint-wise adaptive anchor setting. 
That is, for each joint a certain number of 3D anchors will be adaptively sampled around it in 3D space, in the end-to-end learning way. 
Compared with existing anchor-based methods, which only set fixed static anchors, our anchor setting approach can better exploit joint's local descriptive context. And, the joint's 3D position is jointly determined by all the sampled 3D anchors with ensemble prediction to alleviate the high dependence on accurate 2D HPE result and long sequence input.
Moreover, through the pose-guided feature enhancement module, we embed the 2D pose as spatial cues into visual features to establish a more robust interaction with 3D anchors.

Experimental results on CMU Panoptic~\cite{Joo_2017_TPAMI} and ITOP~\cite{haque2016towards} datasets demonstrate that our proposed weakly supervised 3D HPE method can achieve comparable or even superior results to the current SOTA 3D fully supervised counterparts. 
When large-scale unlabeled RGB-D data (\textit{e.g.}, NTU RGB+D 60~\cite{shahroudy2016ntu}) is used, our approach can even outperform them under the more challenging cross-view test setting. 
The introduction of cross-modality feature enhancement can essentially facilitate performance consistently, especially towards human body's end joints (\textit{hand and foot}). Additionally, our proposed anchor-to-joint based 3D pose lifting method outperforms the SOTA single-frame and most sequence-based methods both under weakly supervised and fully supervised settings on CMU Panopic~\cite{Joo_2017_TPAMI}.
 
Overall, the main contributions of this paper include:
\begin{itemize}
    \item For the first time, we propose the research problem of unified weakly supervised 3D human pose estimation towards RGB image, depth map and point cloud modalities, using the unannotated RGB-D data easily accessible from low-cost RGB-D camera;
    \item UniPose: a novel weakly supervised 3D human pose estimation method that concerns the issues of cross-modality pose prior propogation, weak and self supervision clues exploration, and multi-modality feature enhancement; 
    \item For RGB image and depth map, an anchor-to-joint based 3D pose lifting method is proposed to alleviate the ill-posed 2D-to-3D prediction problem.
\end{itemize}

\section{Related work}\label{sec2:related}

\textbf{Weakly supervised 3D HPE methods.}
The high expense of data collection and annotation for 3D Human Pose Estimation (HPE) encourages the use of unlabeled data under weak supervision. This context introduces two mainstream approaches: multi-view methods and synthetic model-based techniques. Multi-view methods \cite{kocabas2019self,mitra2020multiview} typically require images from three or more views along with complex calibration using camera parameters. These methods face challenges like view-inconsistency and accumulating errors when converting 2D poses into 3D space through triangulation. To mitigate these issues, some works have reduced the number of required views to two \cite{remelli2020lightweight} or attempted to predict camera parameters via networks \cite{wandt2019repnet}. However, acquiring multi-view images remains inherently difficult compared to single-view data.
Some works employ synthetic software \cite{shrivastava2017learning} or renders predicted keypoints onto depth or RGB images using predefined human skeleton models \cite{cai2018weakly}. Such supervision signals from synthetic or rendered images can aid in training 3D pose estimators, though they encounter domain shift problems between synthetic and real data. 
Notably, existing weakly supervised methods mainly focus on the RGB domain with few attention to depth domains such as depth maps and point clouds.

\textbf{RGB based fully-supervised 3D HPE methods.}
With advancements in deep learning and 2D pose estimation~\cite{xiao2018simple,sun2019deep}, the focus has been on lifting predicted 2D poses to 3D, categorized into single-frame and multi-frame methods.
Single-frame methods~\cite{Martinez_2017_ICCV,zhao2023contextaware} emphasize high-quality 2D pose estimation for accurate 3D reconstruction, facing challenges due to the ill-posed nature of 2D-to-3D lifting. Multi-frame methods use temporal information to address these issues, establishing spatial-temporal connections in 2D pose sequences~\cite{pavllo20193d,zhang2022mixste}. 
Despite progress, these methods rely heavily on 2D pose prediction accuracy and are sensitive to errors. In weakly supervised settings, the lack of 3D annotations affects 2D-3D correspondence, causing performance degradation. To mitigate these, our method introduces an anchor-to-joint mechanism and exploits pose priors within 2D estimators, reducing dependency on high-quality annotations and enhancing 3D HPE robustness.

\textbf{Point cloud and depth map based fully-supervised 3D HPE methods.}
Non-deep learning approaches~\cite{he2015depth,shotton2011real} were pivotal in early depth image-based human pose estimation research, utilizing hand-crafted features with regression or classification. Their performance is often inferior to deep learning methods due to limited feature representation.
Recent advancements have seen the emergence of various deep learning-based HPE techniques~\cite{moon2018v2v,xiong2019a2j}, leveraging the powerful fitting capabilities of deep neural networks. Notably, A2J~\cite{xiong2019a2j} introduces an anchor-to-joint mechanism for 3D HPE, enhancing localization accuracy through multi-anchor offsets. Its ensemble prediction approach ensures high performance and generalization. We propose applying this method to solve the 2D-to-3D pose lifting problem by adaptively setting 3D anchors around predicted 2D joints, thus reducing sensitivity to 2D localization errors and improving 3D joint localization.
For point cloud, early works~\cite{obdrvzalek2012real,biswas2019fast,schnurer2019real}, commonly used 2D CNNs to localize human joints and extend them to 3D via depth transformations. However, true 3D representation requires advanced networks like PointNet~\cite{qi2017pointnet}, and PointTransformer~\cite{zhao2021point}, which significantly enhance feature extraction from unordered point clouds and advance 3D HPE~\cite{ge2018hand,ge2018point}. 

\section{Main technical pipeline}\label{sec3:pipeline}

\begin{figure}[t]
  \centering
  \includegraphics[width=0.92\linewidth]{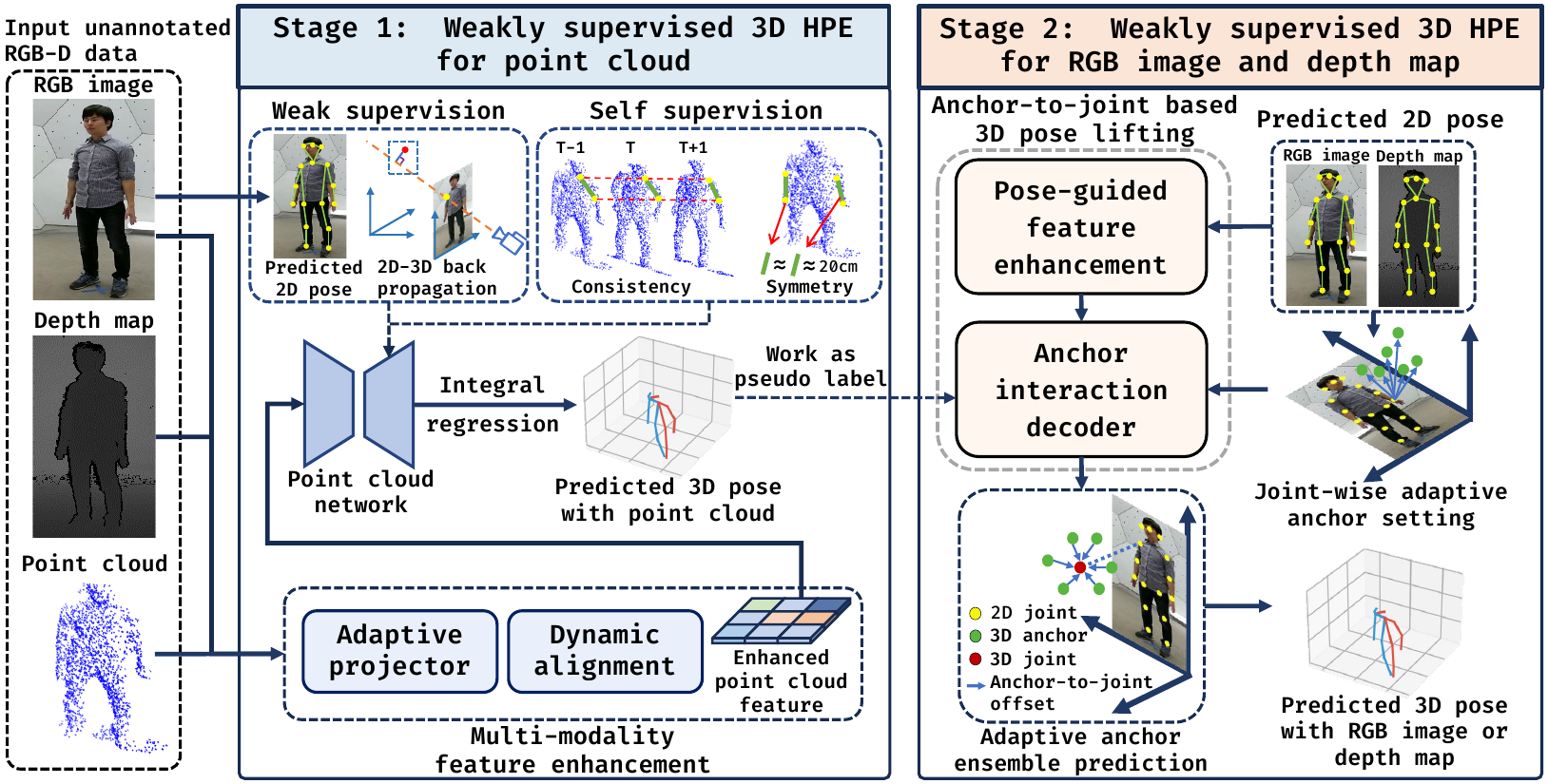}
   \caption{UniPose's main pipeline that runs in a 2-stage way. Within the first stage, weakly supervised 3D HPE is conducted towards point cloud modality under weak and self supervision exploited from unannotated RGB-D data, with multi-modality feature enhancement. Within the second stage, using point cloud's 3D HPE result as 3D pseudo label, weakly supervised 3D HPE is executed to RGB image and depth map. 
    }
   \label{fig:pipeline}
\end{figure}

The main technical pipeline of our proposed weakly-supervised 3D human pose estimation method (\textit{i.e.}, UniPose) towards RGB image, point cloud and depth map uniformly is shown in \cref{fig:pipeline}.
Given the unannotated RGB-D data sequence as input, within the first stage, we perform pose estimation on the point cloud, and obtain 3D human pose through effective self-supervision and weakly-supervision signal extraction, along with multi-modality feature enhancement. 
Within the second stage, adaptive anchor-to-joint based 3D pose lifting is executed to RGB image and depth map for weakly-supervised 3D HPE. 

\section{Weakly supervised 3D HPE for point cloud}

In this section, the proposed weakly supervised 3D HPE method for point cloud will be introduced. We first introduce the construction of supervision signals, including weak supervision from RGB image and self supervision from point cloud. The multi-modality feature enhancement is further introduced to facilitate the quality of predicted 3D pose. 

\subsection{Multi-modality feature enhancement}
\begin{figure}[t]
  \centering
  \includegraphics[width=0.9\linewidth]{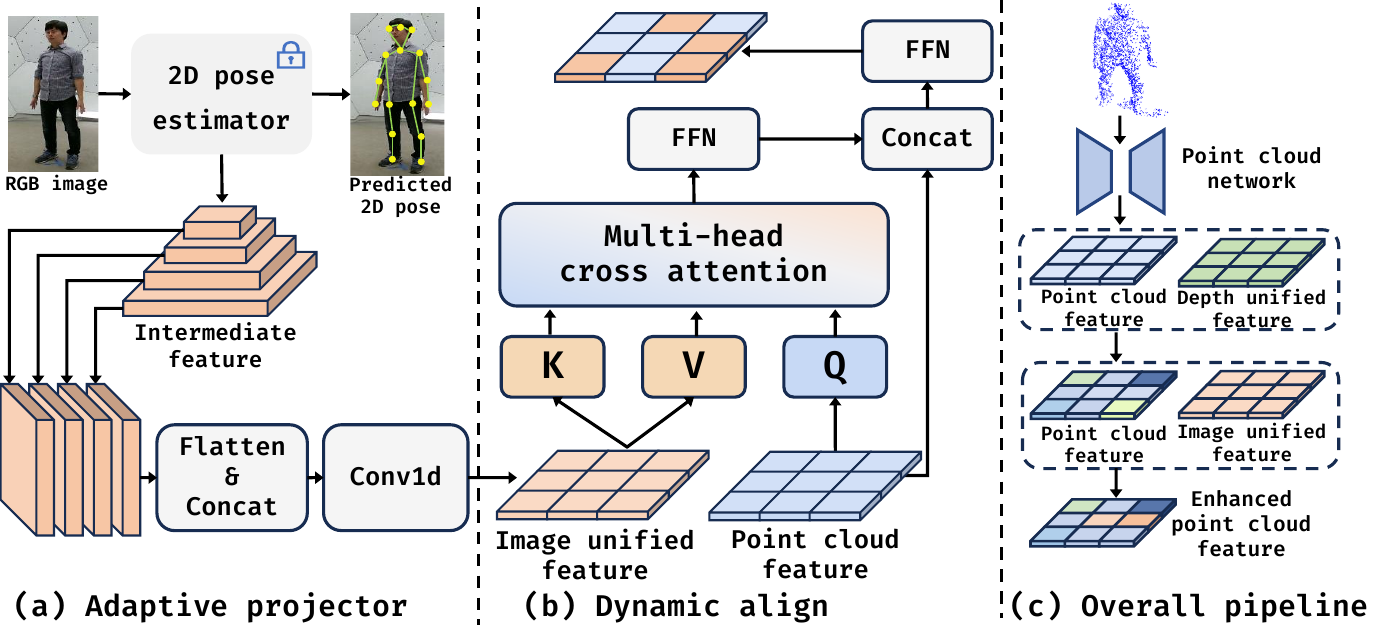}
   \caption{The multi-modality enhancement involves two modules: (a) Adaptive projector and (b) Dynamic align. The depth projector and RGB projector have separate parameters due to the distinct characteristics of RGB image and depth map. Given the homology between point cloud and depth map, point cloud feature will be feature-aligned with depth map and then aligned with RGB features.
    }
   \label{fig:projector}
\end{figure}

UniPose utilizes RGB image $I$ and depth data (depth map 
$D$ and point cloud $P$) as inputs. For point cloud processing, PointNet~\cite{qi2017pointnet} extracts point-level features $F_{point} \in R^{N\times D}$. Regarding images, we leverage the by-products of 2D pose estimators, particularly their intermediate pyramid feature maps, which encode human-oriented spatial context useful for enhancing point cloud representations without extra backbone training. Consequently, we use pretrained HRNet~\cite{sun2019deep} to extract RGB features $F_{rgb}$ and depth map features $F_{depth}$.
Aiming to unify multi-modal features for 3D HPE, our method projects image features into the 3D point feature space for improved depth domain responses. Given a 3D point cloud, only corresponding resolution image pixels can accurately locate points. However, downsampling in the pyramid image backbone results in pyramid feature maps where each pixel corresponds to a subset of the point cloud. Instead of naively flattening these through avgpool, we propose the Adaptive Projector, which linearly regresses pyramid image features layer by layer, concatenates them to fuse detailed and abstract features from different network levels, implicitly encoding spatial information related to joints. Features are then aligned with the point cloud's dimensions ($N \times C$) via a linear layer. Due to the distinct characteristics of RGB images and depth maps, separate parameters are used for Depth and RGB Projectors, with point cloud features first aligned with depth maps, then with RGB features.
After obtaining the features of image $F_{RGB\_uni}\in R^{N\times C}$ and $F_{depth\_uni}\in R^{N\times C}$, in order to better integrate the information from point cloud features with the most relevant image features, we propose the \textbf{Dynamic alignment} strategy (\cref{fig:projector}). Though cross attention mechanism, the network dynamically capture the cross modal correlation. 
A standard point cloud network consists of downsampling encoder and upsampling decoder. Considering the homology between the depth map and the point cloud, we stack the encoder-decoder twice, aligning depth features in the first encoder and RGB features in the second encoder.

\subsection{Weak and self supervision signal}

As aforementioned, an off-the-shelf 2D pose estimator is used to get the 2D pose predictions $C=\{c_k\}_{k=1}^K$ , where K is the number of body joints, and $c_k\in R^2$ is the $k$-th joint on image plane. The point cloud based 3D pose estimator's output is $J = \{j_k\}^K_{k=1}$, and $j_k \in R^3$ is the $k$-th joint coordinate in world coordinate. Now the problem is how to use 2D joint prediction $C \in R^{2\times K}$ to supervise the 3D output $J \in R^{3\times K}$.
The common practice of current weak supervision methods~\cite{chen2019weakly} is to directly project the 3D joint predictions to 2D RGB image plane, distance between the oriented joint locations $\hat{c_k}$ and 2D pose label $c_k$ can be utilized to formulate the loss function as:
\begin{equation}
\mathcal{L}_{rgb}=\sum_{k=1}^K\|c_k-\hat{c}_k\|_2.
\end{equation}

\begin{figure}[t]
    \centering
    \begin{minipage}[b]{0.35\textwidth}
    \includegraphics[width=1.0\linewidth]{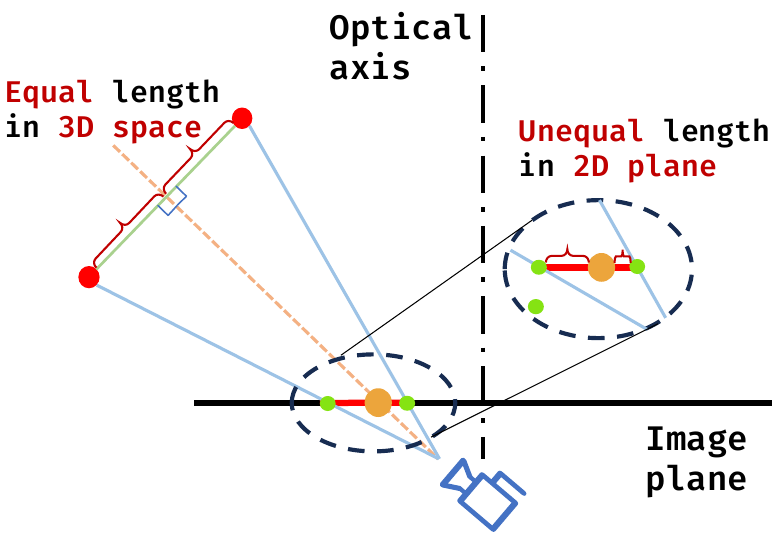}
       \caption{Schematic of 3D-to-2D projection manner. }
       \label{fig:project_2d}       
    \end{minipage}
    \hfill 
    \begin{minipage}[b]{0.64\textwidth}
        \includegraphics[width=1.0\linewidth]{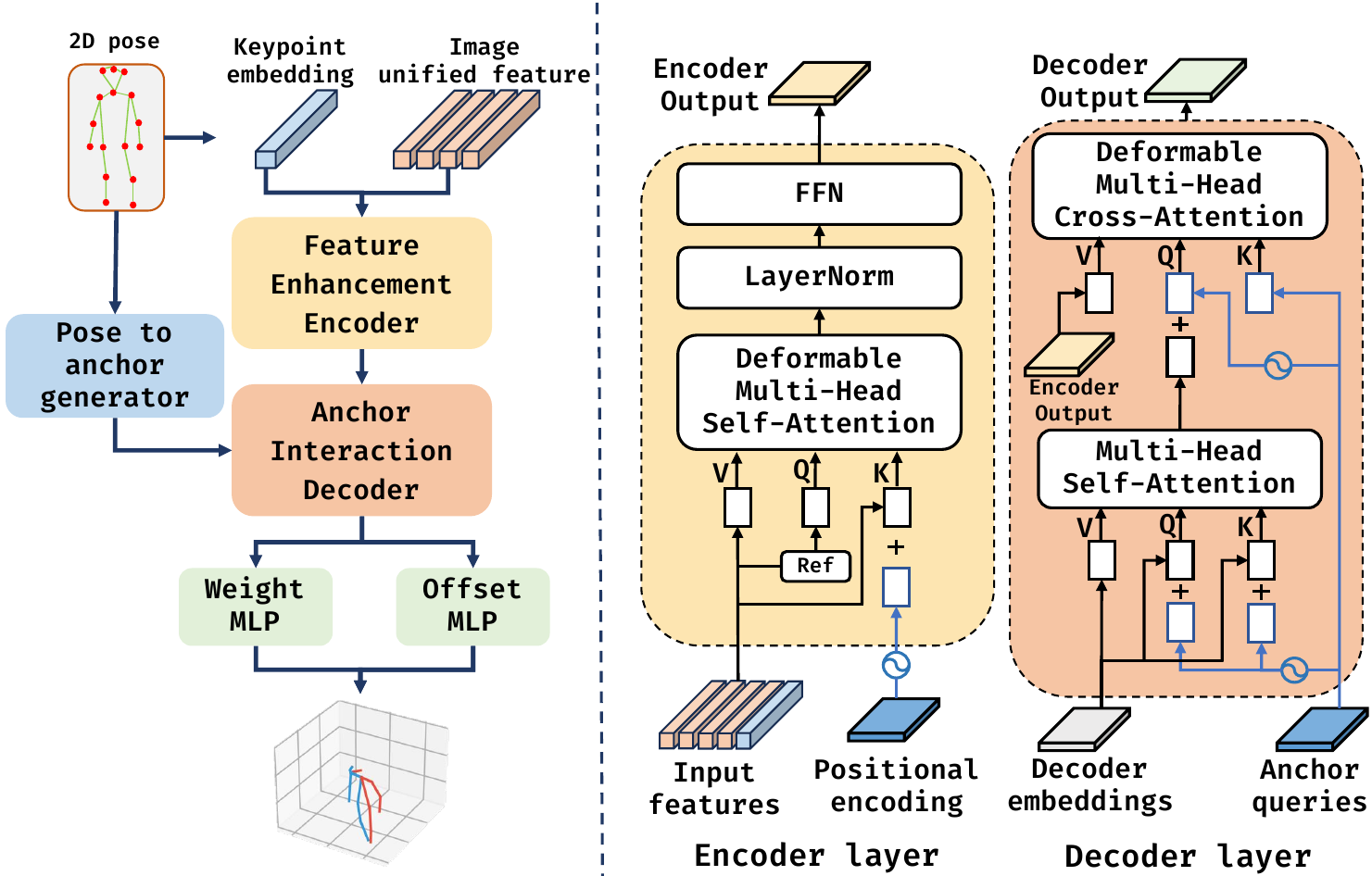}
        \caption{Network structure of stage 2.}
        \label{fig:liftingpipeline}
    \end{minipage}
\end{figure}

We argue that this approach may not be optimal. 
As illustrate in~\cref{fig:project_2d}, when faced with predictions symmetric about the projection rays, the method of projecting 3D joints to the 2D plane causes inconsistent distances of the 3D points in the 2D plane with respect to the 2D joints. Predicted points closer to the optical axis possess smaller errors, while points farther away from the optical axis have larger losses, which affects network's parameter updates in back-propagation and is detrimental to the consistency of weakly-supervised losses in 3D space, ultimately leading to network's tendency to predict joints toward the center of the optical axis.
Accordingly, we propose to back-project the pre-computed keypoints $c_k$ on image plane to ray forms in 3D space. Subsequently, the distance between predicted 3D points and back-projected 3D rays is computed. This proposed back-projection method can yield more robust 3D training supervision signal.
Technically, the distance between $j_k$ and the ray back-projected by $c_k$ needs to be calculated. By back-projecting $c_k$ into a ray in 3D space using the camera's intrinsic parameters, and subsequently calculating the distance from point cloud network's predicted 3D joint $\hat{j_k}$ to corresponding ray. We leverage this distance metric $\mathcal{D}_{p-ray,k}$ to redefine the weakly-supervised loss function, thereby yielding a more spatially-consistent weakly supervised loss:
The weakly supervised signal from 2D joints is then formulated as:

\begin{equation}
    \mathcal{L}_{2d} = \sum_{k=1}^K\frac{\mathcal{D}_{p-ray,k}}{\mu}
\end{equation}

where we set a hyper parameter $\mu = \left\|\hat{j_k}\right\|_2$ to prevent the network from converging to trivial solution(i.e., zero vector).

To further exploit the intrinsic prior of human pose in depth domain, we follow~\cite{wu2022c3p} to build self-supervision signals with the constraints on human body symmetry $\mathcal{L}_{sym}$, bone average length $\mathcal{L}_{bone}$ and pose temporal motion consistency $\mathcal{L}_{con}$.

Overall, the total loss for point cloud branch is formulated as:
\begin{equation}
    \mathcal{L} = \lambda_1\mathcal{L}_{2d} + \lambda_2\mathcal{L}_{bone} + \lambda_3\mathcal{L}_{sym} + \lambda_4\mathcal{L}_{con}
\end{equation}
here we set $\lambda_1 = 10$,$\lambda_2 = 1$ ,$\lambda_3 = \lambda_4 =0.1$ to balance each loss item.

\section{Weakly supervised 3D HPE for RGB and depth map}


\subsection{Cross modal feature enhancement}
The architecture of proposed pipeline is shown in~\cref{fig:liftingpipeline}. We directly use the unified feature representation $F_{img\_uni}$ and predicted 2D pose $C \in R^{K\times2}$ as input for the pose lifting network. To enhance the leverage of pose prior,  we propose to treat the predicted 2D pose as guiding information that encoding spatial descriptive clues. Specifically, we encode the 2D pose into the same number of channels as the input features through a linear layer and concatenate them together to serve as the fused input features:
\begin{equation}
   F_{fuse} = Concat(F_{img\_uni}, Emb_{pose}) 
\end{equation}
\begin{equation}
  PE_{fuse} = Concat(PE_{img\_uni}, PE_{pose}) 
\end{equation}
where $PE$ indicates positional embedding~\cite{zhu2020deformable}.
Given fused feature, we extract informative spatial context through deformable attention~\cite{zhu2020deformable}. 
The addition of positional embedding and pose embedding provide stronger relative positional deviations to facilitate feature enhancement.
\begin{equation}
   Q = F_{fuse} + PE_{fuse}, \quad K = ref(F_{fuse}), \quad V = F_{fuse}
\end{equation}

\subsection{Pose-guided anchor-to-joint prediction}


Current work commonly predicts 3D poses through a point-to-point mapping of 2D poses directly to 3D poses, which makes these methods generally more sensitive to the noise in predicting 2D poses. In this work, we further utilize pose prior and propose the 2D pose to anchor to joint mechanism.
For the input image, we will adaptively sample $N_{local}$ anchors in the 3D space for each joint. The offsets of 2D joint relative to sampled anchors are learnable. This ensures that anchors are distributed around the joints as much as possible, resulting in more localized prediction results.
At the same time, for the input image size of $256\times 192$,  $16 \times 12$ global fix anchors are evenly distributed at depth $0$ to capture global image context and maintain generalization. The overall anchors consist of a combination of local anchors and global anchors as $ A = \sum_{j\in J}A_{local}(j) + A_{global}$ and the total numbers of anchor is $N_{anchor} = N_{local} \times K + N_{global}$ , where we set $N_{local} = 20$ and $N_{global} = 192$.
After generating pose-guided anchors $A$, we establish the interaction between anchors and enhanced spatial features through deformable cross-attention. The input can be formulated as:
\begin{equation}
    Q = F_{dec} + Emb_A, \quad K = A, \quad V = F_{enc}
\end{equation}
where $F_{dec}$ and $F_{enc}$ indicates the features output by decoder layer and feature-enhanced encoder.  Here we serve anchors (i.e. $K = A$ ) as reference points. 
Then, the output anchor-related features are fed into the weight estimator and offset estimator to estimate the weight $W(a,j)\in R^{N_{anchor} \times K}$ and offset $O(a,j) \in R^{N_{anchor} \times K \times 3}$ of each anchor to relative joints. The positions of each joint can be calculated as the weighted sum of predicted results of the total anchor set:

\begin{equation}
    j = \sum W(a,j) \times (a + O(a, j)), \quad a \in A_{local}(j) \cup A_{global}
\end{equation}

\section{Experiment}\label{sec4:experiment}

\begin{table*}[t]
\tiny
\centering

\caption{Performance on CMU Panoptic Dataset~\cite{Joo_2017_TPAMI} for point cloud input. P2P and PT mean our method use different backbone network. The unit of MPJPE is cm.}
\setlength{\tabcolsep}{3pt}
\begin{tabular}{lccccccccccc}
\bottomrule
\multicolumn{1}{l}{} &
  \multicolumn{1}{l}{Nose} &
  \multicolumn{1}{l}{Eyes} &
  \multicolumn{1}{l}{Ears} &
  \multicolumn{1}{l}{Shoulders} &
  \multicolumn{1}{l}{Elbows} &
  \multicolumn{1}{l}{Wrists} &
  \multicolumn{1}{l}{Hips} &
  \multicolumn{1}{l}{Knees} &
  \multicolumn{1}{l}{Ankles} &
  \multicolumn{1}{l}{mAP} &
  \multicolumn{1}{l}{MPJPE} \\ \hline
\multicolumn{11}{c}{Fully-supervised methods}                                                    \\ \hline
\multicolumn{1}{l|}{HandPoint\cite{ge2018hand}}       & 79.8 & 78.9 & 79.6 & 80.6 & 5.3 & 0.2 & 89.6 & 84.5 & 75.1 & 62.8 & 12.0\\
\multicolumn{1}{l|}{P2P\cite{ge2018point}}      & 98.1 & 98.1 & 98.2 & 96.9 & 95.0 & 89.8 & 94.5 & 94.1 & 93.7 & 95.2 & 4.1 \\
\multicolumn{1}{l|}{PT\cite{zhao2021point}}       &  \textbf{99.6} & \textbf{99.3} & \textbf{99.2} & \textbf{98.8} & \textbf{97.0} & \textbf{92.0} & \textbf{96.8} & \textbf{95.0} & \textbf{95.7} & \textbf{96.9} & \textbf{3.3} \\
\hline
\multicolumn{11}{c}{Weakly-supervised methods} \\                                  \hline
\multicolumn{1}{l|}{C3P(P2P)\cite{wu2022c3p}} & 96.3 & 95.8 & 95.0 & 93.9 & 91.4 & 81.5 & 90.9 & 78.4 & 85.2 & 89.4 & 6.1 \\ 
\multicolumn{1}{l|}{C3P(PT)\cite{wu2022c3p}} & 99.1 & 98.6 & 95.9 & 95.4 & 94.4 & 85.4 & 93.1 & 91.1 & 94.0 & 93.8 & 5.3 \\ 
\rowcolor{gray!20}
\multicolumn{1}{l|}{UniPose(P2P)} & 98.9 & 98.8 & 95.3 & 94.9 & 95.1 & 85.3 & 92.9 & 87.4 & 92.2 & 93.7 & 5.2 \\ 
\rowcolor{gray!20}
\multicolumn{1}{l|}{UniPose(PT)} & \textbf{99.2} & \textbf{98.8} & \textbf{96.1} & \textbf{95.5} & \textbf{94.6} & \textbf{85.7} & \textbf{93.1} &\textbf{ 91.4} & \textbf{94.3 }& \textbf{95.3} & \textbf{4.8} \\ 
\bottomrule
\end{tabular}
\vspace{5pt}
\caption{Performance on CMU Panoptic Dataset~\cite{Joo_2017_TPAMI} for image input. }
\begin{tabular}{lccccccccccc}
\bottomrule
\multicolumn{1}{l}{} &
  \multicolumn{1}{l}{Nose} &
  \multicolumn{1}{l}{Eyes} &
  \multicolumn{1}{l}{Ears} &
  \multicolumn{1}{l}{Shoulders} &
  \multicolumn{1}{l}{Elbows} &
  \multicolumn{1}{l}{Wrists} &
  \multicolumn{1}{l}{Hips} &
  \multicolumn{1}{l}{Knees} &
  \multicolumn{1}{l}{Ankles} &
  \multicolumn{1}{l}{mAP} &
  \multicolumn{1}{l}{MPJPE} \\ \hline
\multicolumn{11}{c}{Fully-supervised methods (Using GT 3D pose label)}                                                    \\ \hline
\multicolumn{1}{l|}{PoseFormerV2\cite{zhao2023poseformerv2} }  & 94.3 & 94.7 & 94.2 & 92.2 & 88.3 & 81.4 & 88.5 & 82.9 & 81.7 & 85.6 & 7.31 \\
\multicolumn{1}{l|}{A2J-Trans\cite{jiang2023a2j}}       & 95.3 & 95.1 & 94.2 & 92.9 & 89.1 & 81.8 & 89.5 & 83.1 & 83.7 & 87.1 & 5.53\\
\multicolumn{1}{l|}{CA-PF\cite{zhao2023contextaware}}       &  98.1 & 98.2 & 98.1 & 96.8 & 96.0 & 92.0 & 96.4 & 94.0 & 94.8 & 94.7 & 3.83 \\
\rowcolor{gray!20}
\multicolumn{1}{l|}{UniPose}       &  \textbf{99.3} & \textbf{98.3} & \textbf{98.7} & \textbf{97.0} & \textbf{97.1} & \textbf{92.9} & \textbf{96.5} & \textbf{94.4} & \textbf{95.2} & \textbf{95.1} & \textbf{3.77} \\
\hline
\multicolumn{11}{c}{Weakly-supervised methods (Using predicted 3D pose pseudo label)} \\                                  \hline
\multicolumn{1}{l|}{PoseFormerV2\cite{zhao2023poseformerv2} }      & 93.2 & 93.9 & 93.0 & 90.0 & 85.8 & 78.6 & 87.7 & 80.1 & 80.8 & 84.1 & 8.23 \\
\multicolumn{1}{l|}{A2J-Trans\cite{jiang2023a2j}}       & 94.1 & 94.3 & 93.2 & 90.9 & 88.7 & 80.2 & 88.3 & 80.8 & 80.9 & 84.8 & 6.05\\
\multicolumn{1}{l|}{CA-PF\cite{zhao2023contextaware}}       &  95.8 & 96.1 & 95.9 & 93.1 & 90.1 & 82.0 & 90.1 & 83.6 & 84.8 & 86.8 & 5.85 \\
\rowcolor{gray!20}
\multicolumn{1}{l|}{UniPose}       &  \textbf{96.4} & \textbf{97.4} & \textbf{96.2} & \textbf{93.9} & \textbf{93.2} & \textbf{83.9} & \textbf{92.5} & \textbf{84.2} & \textbf{86.9} & \textbf{89.9} & \textbf{5.25} \\
\bottomrule
\end{tabular}
\label{tab:CMU}
\end{table*}

\subsection{Datasets and evaluation metrics}

For point cloud, we use \textbf{CMU Panoptic}~\cite{Joo_2017_TPAMI} and \textbf{ITOP}~\cite{haque2016towards} to validate our point cloud network. \textbf{CMU Panoptic} contains corresponding RGB image and \textbf{ITOP Dataset} contains corresponding depth image, we use them to further validate the effectiveness of the image network.
We conduct the cross-dataset test on \textbf{NTU RGB+D} dataset to demonstrate that, trained on large-scale unannotated dataset can even better adapt to scene variation than the fully-supervised counterparts.
The evaluation metrics is the mean average precision (mAP) with 10-cm rule. The MPJPE is also used as evaluation metric. Details about datasets are provided in supplementary material.

\subsection{Implementation details}

UniPose is implemented using PyTorch. For point cloud branch, the input point number is set to $2048$, and point cloud normalization operations if the same as P2P~\cite{ge2018point}. For image branch, we directly crop the images and resize them to $256 \times 256$ resolution. We train our model using AdamW optimizer and the learning rate is set to $1\times 10^{-4}$ with a weight decay of  $1\times 10^{-4}$ in all cases. 

\subsection{Comparison with fully-supervised method}


\textbf{CMU Panoptic}~\cite{Joo_2017_TPAMI}
For point cloud, UniPose is compared with the fully supervised methods (HandPoint\cite{ge2018hand}, P2P\cite{ge2018point},PT\cite{zhao2021point}). 
UniPose achieves comparable results to fully-supervised counterparts. Without labeled data, the result os only inferior less than 2\% in mAP and about 1cm in MPJPE. Meanwhile, UniPose can be applied to the different point cloud networks (P2P~\cite{ge2018point} and PT~\cite{zhao2021point}. This indicates that the proposed weakly-supervised 3D HPE manner towards point cloud is not sensitive to the choice of backbone network either CNN or Transformer.
For RGB image input, we investigate the effectiveness of proposed image network under fully-supervised and weakly-supervised conditions. When using GT 3D pose label, our method outperforms SOTA method~\cite{zhao2023contextaware}. The advantage of our method is even greater when training with noisy labels, which shows the good tolerance of our method to noise.
    
    

\textbf{ITOP dataset}~\cite{haque2016towards}.
ITOP dataset only contains depth images. According, we use the 2D annotation on depth image to replace the supervision signal from RGB image~\cite{wu2022c3p}. SOTA fully supervised methods are compared to verify UniPose's effectiveness. The performance comparison results are listed in \Cref{tab:ITOP}. It can be observed that UniPose can achieve comparable results to fully-supervise methods. With the same 3D network, UniPose's performance drop is slight (less than 1.2\% in mean error).

    
    

\begin{table}[t]
\scriptsize
\centering
\caption{Results on ITOP Dataset~\cite{haque2016towards}. (P2P) and (PT) means our method use different 3D point cloud-based network P2P~\cite{ge2018point} and PT~\cite{zhao2021point}.}
\setlength{\tabcolsep}{4pt}
\begin{tabular}{lcccccccccc}
\bottomrule

\multicolumn{1}{l}{} &
  \multicolumn{1}{l}{Head} &
  \multicolumn{1}{l}{Neck} &
  \multicolumn{1}{l}{Shoulders} &
  \multicolumn{1}{l}{Elbows} &
  \multicolumn{1}{l}{Hands} &
  \multicolumn{1}{l}{Torso} &
  \multicolumn{1}{l}{Hips} &
  \multicolumn{1}{l}{Knees} &
  \multicolumn{1}{l}{Feet} &
  \multicolumn{1}{l}{mean} \\ \hline
\multicolumn{11}{c}{Fully-supervised methods}                                                    \\ \hline
\multicolumn{1}{l|}{RF\cite{shotton2011real}}          & 63.8 & 86.4 & 83.3 & 73.2 & 51.3 & 65.0 & 50.8 & 65.7 & 61.3 & 65.8 \\
\multicolumn{1}{l|}{VI\cite{haque2016towards}}          & 98.1 & 97.5 & 96.5 & 73.3 & 68.7 & 85.6 & 72.0 & 69.0 & 60.8 & 77.4 \\
\multicolumn{1}{l|}{CMB\cite{wang2018convolutional}}         & 97.7 & 98.5 & 75.9 & 62.7 & \textbf{84.4} & 96.0 & 87.9 & 84.4 & 83.8 & 83.3 \\
\multicolumn{1}{l|}{REN-9*6*6\cite{guo2017towards}}   & \textbf{98.7} & \textbf{99.4} & 96.1 & 74.7 & 55.2 & \textbf{98.7} & 91.8 & 89.0 & 81.1 & 84.9 \\
\multicolumn{1}{l|}{V2V*\cite{moon2018v2v}}        & 98.3 & 99.1 & \textbf{99.2} & \textbf{80.4} & 67.3 & \textbf{98.7} & \textbf{93.2} & \textbf{91.8} & \textbf{87.6} & \textbf{88.7} \\
\multicolumn{1}{l|}{A2J\cite{xiong2019a2j}}         & 98.5 & 99.2 & 96.2 & 78.9 & 68.4 & 98.5 & 90.9 & 90.8 & 86.9 & 88.0 \\
\multicolumn{1}{l|}{P2P\cite{ge2018point}}       &  97.6 & 98.6 & 95.5 & 78.0 & 63.9 & 97.8 & 88.9 & 90.6 & 85.1 & 86.5 \\
\multicolumn{1}{l|}{{PT}\cite{zhao2021point}}       &  97.7 & 98.6 & 96.0 & 78.6 & 64.1 & 98.5 & 90.1 & 90.8 & 86.5 & 87.2 \\
\hline
\multicolumn{11}{c}{Weakly-supervised method}                                                          \\ \hline
\multicolumn{1}{l|}{C3P(P2P)\cite{wu2022c3p}} & 97.0 & 97.4 & 91.4 & 75.6 & 63.1 & 96.4 & 85.2 & 88.7 & 84.2 & 84.5 \\ 
\multicolumn{1}{l|}{C3P(PT)\cite{wu2022c3p}} & 97.3 & 97.0 & 91.3 & 76.3 & 66.5 & 94.9 & 81.7 & 87.8 & 83.4 & 84.3 \\ 
\rowcolor{gray!20}
\multicolumn{1}{l|}{UniPose(P2P)} & 97.2 & \textbf{97.4} & 91.7 & 75.7 & 66.7 & 96.5 & 86.4 & \textbf{89.9} & \textbf{86.8} & \textbf{86.1} \\ 
\rowcolor{gray!20}
\multicolumn{1}{l|}{UniPose(PT)} & \textbf{97.6} & 97.3 & \textbf{92.1} & \textbf{76.8} & \textbf{66.8} & 95.6 & 84.7 & 89.5 & 86.6 & 86.0 \\ 
\bottomrule
\end{tabular}
\label{tab:ITOP}
\end{table}

 \textbf{Cross-view test.} We evaluate the generalization capabilities of different HPE methods through cross-view testing on CMU Panoptic, where each method is trained on one view and tested on another. Unlike traditional methods that rely on complex data augmentation strategies, UniPose enhances its versatility by leveraging large-scale unlabeled RGB-D data (NTU RGB+D dataset~\cite{shahroudy2016ntu}), demonstrating superior performance and potential for further improvement in practical applications. Experimental results show that, although the performance of all 3D HPE methods declines under challenging cross-view tests, UniPose achieves better performance than fully supervised methods by utilizing large-scale unlabeled data. This indicates that UniPose can benefit from easily collectable large-scale unlabeled data, making it suitable for practical applications.
 

    

\begin{figure}[t]
	\centering
	\begin{minipage}{0.49\linewidth}
		\centering
		\includegraphics[width=0.9\linewidth]{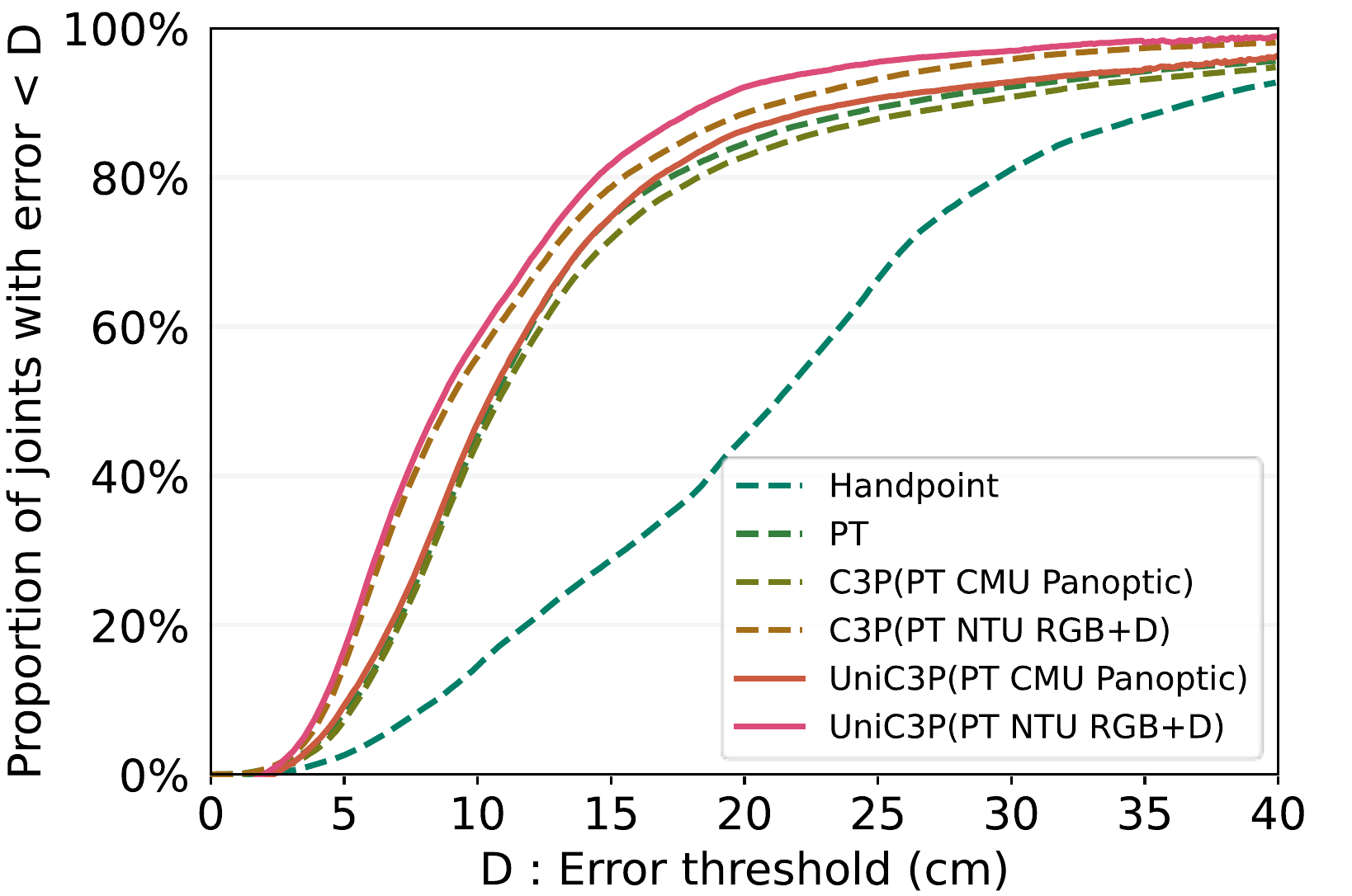}
		\caption{Precision with different D.}
		\label{fig_ntu1}
	\end{minipage}
	\begin{minipage}{0.49\linewidth}
		\centering
		\includegraphics[width=0.9\linewidth]{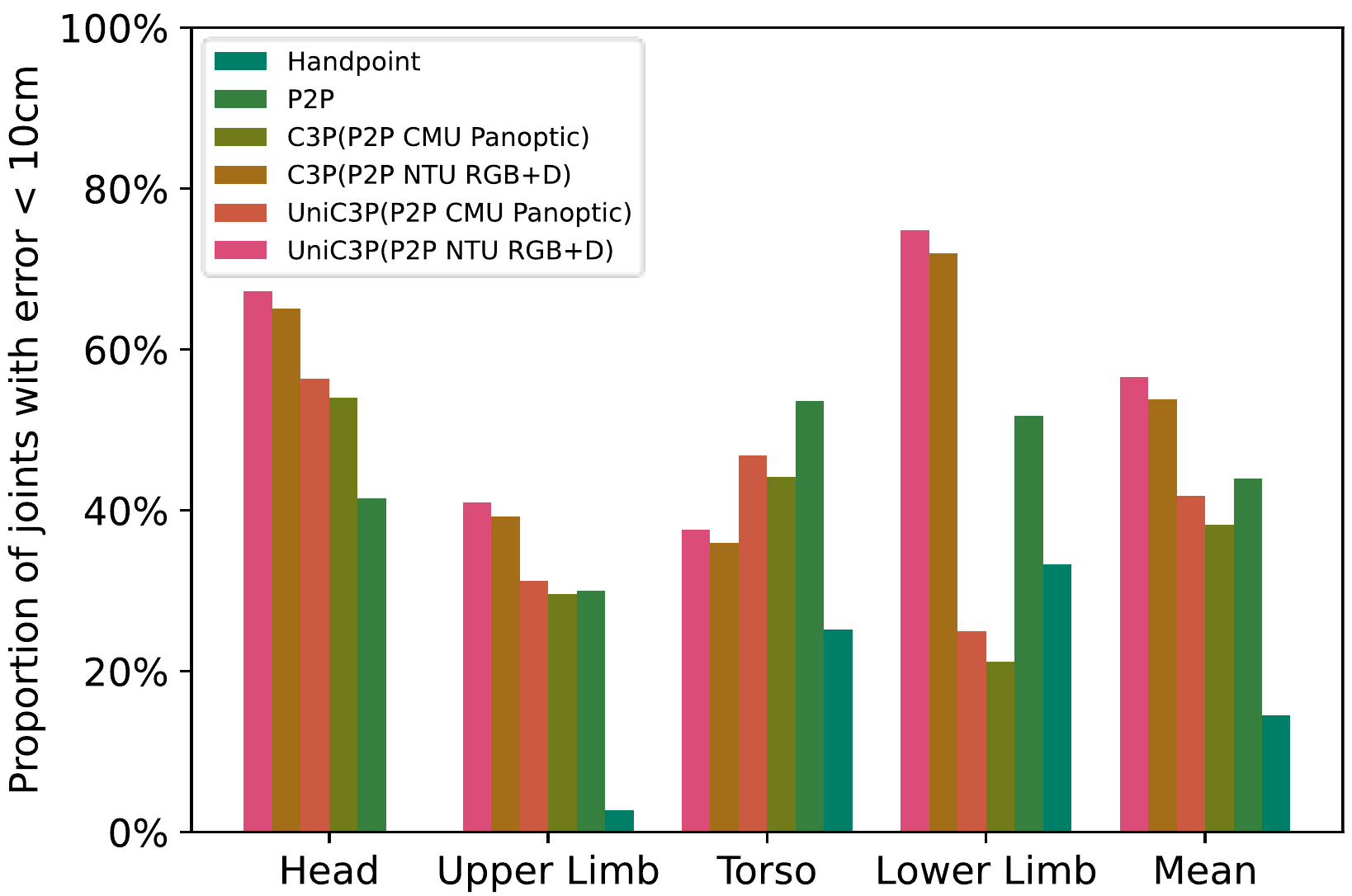}
		\caption{Joints' mean precision.}
		\label{fig_ntu2}
	\end{minipage}
\end{figure}

\subsection{Ablation study}
We performed comprehensive ablation studies to validate our propositions, focusing on the effectiveness of \textbf{weak and self supervision signals} in point cloud networks, \textbf{key components' effectiveness} in image networks, and the \textbf{impact of pseudo 3D pose label quality} from point cloud networks on image network performance. Details are provided in the supplementary material.
\section{Conclusion}
We propose UniPose, a unified cross-modality pose prior propagation method for weakly supervised 3D human pose estimation using unannotated single-view RGB-D sequences (RGB, depth, and point cloud). 
UniPose transfers 2D HPE annotations from large-scale RGB datasets to 3D domain via self-supervised learning on easily acquired RGB-D sequences, eliminating the need for labor-intensive 3D keypoint annotations. 
Leveraging  2D pose prior as weak supervision with a physical prior as self supervision, UniPose achieves high-quality 3D pose estimation through multi-modality enhancement.
By treating point cloud network's 3D HPE results as pseudo ground truth, our anchor-to-joint prediction method performs 3D lifting on RGB and depth networks. 
Experiments show that UniPose achieves comparable performance to fully supervised methods. 
%
%
%
\subsubsection*{Acknowledgment.}
This work is supported by the National Natural Science Foundation of China (Grant No. 62271221).
\bibliographystyle{splncs04}
\bibliography{main}

\newpage
\section{Supplementary material}
\subsection{Datasets}

\textbf{CMU Panoptic Dataset}~\cite{Joo_2017_TPAMI} contains 480 VGA videos, 31 HD videos and 10 Kinect videos, including depth and RGB images in indoor scene. Calibration and synchronous data are also provided. The 3D human pose annotation for VGA and HD videos are provided at the same time. We use 3D human pose in HD videos, the calibration clue, and synchronous data to get 3D human pose labels for Kinect. Since we only focus on point cloud based 3D human pose estimation for single person, experiments are carried on the "range of motion" class Kinect data with 9 available video sequences. The 6th and 9th videos ("171204 pose6, 171206 pose3") are used for testing, and the remaining 7 videos are for training.

\textbf{ITOP Dataset}~\cite{haque2016towards} is a widely used benchmark dataset in depth image based and point cloud based 3D human pose estimation. It contains 40K training and 10K testing depth images and point clouds data for each front-view and top-view track. This dataset contains 20 actors and 15 human body parts are labeled with 3D coordinates relative to the depth camera. In our experiment, we use the front-view data to evaluate the effectiveness of the method. 

\textbf{NTU RGB+D Dataset}~\cite{shahroudy2016ntu} is a large-scale RGB-D action recognition dataset. It contains over 40 subjects and 60 actions. The actions can cover most daily behavior. There are 17 different scenes in it. The size and diversity are much larger than current human pose dataset. This dataset also contains 3D skeleton joint position labeled by Kinect V2 SDK. But the annotation accuracy is not satisfactory. Nevertheless, the weakly-supervised setting of our method can well leverage this large-scale RGB-D dataset. We conduct the cross-dataset test to demonstrate that, UniPose trained on large-scale unannotated dataset can even better adapt to scene variation than the fully-supervised counterparts.

\subsection{Ablation study}

\subsubsection{Effectiveness of point cloud branch}
 
\textbf{Weak supervision signal}. To verify the effectiveness of the proposed weak supervision information via 2D-to-3D ray projection against existing 3D-to-2D projection way, they are compare on ITOP dataset. 2D GT human pose annotation is used to resist the effect of 2D HPE. The results are listed in~\cref{tab:ablation_ray}. 
The 2D-to-3D ray projection is superior to existing 3D-to-2D counterpart, which can better measure the distance between the predicted and target joints in world coordinate system. For joints (\textit{e.g.}, feet and hands) of high freedom degrees, the 3D-to-2D ray projection supervision outperforms the 3D-to-2D projection strategy by large margins (\textit{i.e.}, +7.7\% on feet and +14.5\% on hands) at mAP.

\begin{table*}
\scriptsize
\centering
\caption{Performance comparison between different weak supervision methods (\textit{i.e.}, our 2D-to-3D ray projection manner vs. 3D-to-2D projection way) on ITOP dataset.}
\vspace{-2ex}
\setlength{\tabcolsep}{1.2mm}{
\begin{tabular}{ccccccccccc}
\bottomrule  & \multicolumn{10}{c}{mAP}    \\ \hline
 & Head & Neck & Shoulders & Elbows & Hands & Torso & Hips & Knees & Feet & mean \\ \hline
\multicolumn{1}{l|}{3D-to-2D manner}   & 95.9 & 95.8 & 90.4 & 68.2 & 49.9 & 91.8 & 81.3 & 85.9 & 78.2 & 79.2 \\ \cline{1-1}
\multicolumn{1}{l|}{2D-to-3D manner (ours)} & \textbf{97.2} & \textbf{97.4} & \textbf{91.7} & \textbf{75.7} & \textbf{66.7} & \textbf{96.5} & \textbf{86.4} & \textbf{89.9} & \textbf{86.8} & \textbf{86.1} \\ \bottomrule
\end{tabular}}

\label{tab:ablation_ray}
\end{table*}

\textbf{Self supervision signal}.
We conduct an ablation experiment on self supervision signals based on physical priors. As shown in~\cref{tab:ablation_selfsupervision}, spatial-temporal supervision signals based on human physics play a relatively important role in prediction quality.
\begin{table}
\footnotesize
\centering
\caption{Effectiveness of self supervision signals within UniPose.
}
\vspace{-2ex}
\setlength{\tabcolsep}{5mm}{
\begin{tabular}{ccc}
\bottomrule
Component       & Mean error(cm) & mAP(@10cm) \\ \hline
w/o bone length $\mathcal{L}_{len}$ & 5.87      &  89.8   \\
w/o symmetry $\mathcal{L}_{sym}$    & 5.62       &  91.2   \\
w/o consistency $\mathcal{L}_{con}$ & 6.14       &  88.1   \\
UniPose (ours)& \textbf{5.23}&  \textbf{93.7}\\ 
\bottomrule
\end{tabular}}
\label{tab:ablation_selfsupervision}
\end{table}

\subsubsection{Effectiveness of image branch}
 
\textbf{Investigation on the quality of 3D pose pseudo label}. 
\cref{tab:ablation_label} presents an analysis of the quality of 3D pose pseudo labels generated by different methods, comparing their performance against ground truth (GT) annotations. The results indicate that both Point-to-Point (P2P)~\cite{ge2018point} output and Point Transformer (PT)~\cite{zhao2021point} output generate high-quality pseudo labels, with PT output showing a slight edge over P2P output.
Nonetheless, the comparison with GT annotations underscores the need for continued advancements in generating high-fidelity 3D pose pseudo labels to bridge the remaining performance gap.

\begin{table}
    \centering
    \scriptsize
\caption{Investigation on the quality of 3D pose pseudo label.}
\label{tab:ablation_label}
\setlength{\tabcolsep}{1.2mm}{
    \begin{tabular}{l|cccccccccccc}
    \toprule
        Label type & Nose & Eyes & Ears & Shoulders  & Elbows & Wrists & Hips & Knees & Ankles & mAP & MPJPE \\ \hline
        P2P output & 96.9 & 94.6 & 93.7 & 92.3 & 91.2 & 74.6 & 89.2 & 82.9 & 87.1 & 88.1 &  5.81 \\
        PT output & 97.2 & 94.6 & 95.1 & 93.1 & 91.8 & 76.2 & 90.3 & 84.5 & 87.5 & 89.9 & 5.25 \\
        GT & 97.5 & 95.9 & 95.7 & 94.7 & 93.8 & 83.9 & 92.9 & 84.6 & 88.7 & 95.1 & 3.77 \\
    \bottomrule
    \end{tabular}}
\end{table}

\textbf{Key components of image networks}. 
To investigate the contributions of key components in image networks, we conducted ablation studies as summarized in~\cref{tab:RGB_component}.
The results indicate that each component plays a significant role in enhancing the overall performance of the image network. Each component contributes uniquely to refining the feature representations, capturing discriminative keypoint features, and accurately generating anchor points, collectively driving the network towards superior performance. 
Especially, when the pose-to-anchor generator is added, the performance reaches its peak, with an mAP of 89.9 and an MPJPE of 5.34. This component appears to be crucial for generating accurate anchors based on the 2D pose priors, which in turn enhances the overall alignment and positioning of the joints.
\begin{table}
    \centering
    \scriptsize
\caption{Ablations on key components of image network.}
\label{tab:RGB_component}
\setlength{\tabcolsep}{1.2mm}{
    \begin{tabular}{ccccc}
    \toprule
        Finetuned projector & Keypoint Embedding & Pose to Anchor Generator & mAP & MPJPE  \\ \hline
         &  &  & 84.8 & 6.06\\
        $\checkmark$ &  &  & 85.4 & 5.91 \\
        $\checkmark$ & $\checkmark$ & & 86.7 & 5.73  \\
        $\checkmark$ & $\checkmark$ & $\checkmark$ & 89.9 & 5.25 \\
    \bottomrule
    \end{tabular}}
\end{table}
\end{document}